\documentclass[11pt]{article}
\usepackage[
    left=2.54cm,
    right=2.54cm,
    top=2.54cm,
    bottom=2.54cm
]{geometry}

\usepackage{tikz}

\usepackage[utf8]{inputenc} 
\usepackage[T1]{fontenc}    
\usepackage{hyperref}       
\usepackage{url}            
\usepackage{booktabs}       
\usepackage{amsfonts}       
\usepackage{amsmath,amsthm,amssymb}
\usepackage{nicefrac}       
\usepackage{microtype}      
\usepackage{xcolor,graphicx}         
\usepackage{capt-of}        
\usepackage{ifthen}
\usepackage{hyperref}
\usepackage{url}
\usepackage{natbib}

\definecolor{darkred}{rgb}{0.8,.1,0.1}
\definecolor{darkgreen}{rgb}{0.1,0.8,0.1}
\definecolor{darkblue}{rgb}{0.1,0.1,0.8}

\newtheorem{theorem}{Theorem}
\newtheorem{remark}[theorem]{Remark}

\newcommand{\yes}{{\color{darkgreen}\checkmark}}
\newcommand{\no}{\color{darkred}{\Large$\times$}}

\newcommand{\turnrow}[1]{%
\raisebox{.65cm}{#1} &
\includegraphics[width=0.15\linewidth]{turn_#1_full_view_pre_llm.png} &
\includegraphics[width=0.15\linewidth]{turn_#1_obs_view_for_llm.png} &
\includegraphics[width=0.07\linewidth]{turn_#1_zoom_view_for_llm.png} &
\includegraphics[width=0.15\linewidth]{turn_#1_full_view_post_llm.png}\\
}

\author{Gabriel~Turinici \\ 
CEREMADE CNRS\\
  Université Paris Dauphine - PSL\\
  Paris 75116 FRANCE\\
  \texttt{gabriel.turinici@dauphine.fr} \\ \url{https://turinici.com}
}
\title{Spatial Strategies, Not Actions: Vector-Quantized Geodesics as Tools for LLM-Driven Agents}

\begin{document}
\maketitle

\begin{abstract}
Large language model (LLM)
 based agents are often criticized for lacking spatial understanding and mainly exploiting statistical text patterns. We investigate their spatial comprehension through an architecture combining geometrical tools with a LLM  serving as a high-level orchestrator in grid-world environments. The agent first collects geodesic trajectories, which are then vector-quantized to extract a representative subset. 
Offline, the LLM associates a natural language description of the underlying behavioral patterns to each selected trajectory, making it a tool.
Online, the LLM chooses the appropriate tool conditioned on the current state and goal. Low-level control is handled by primitive actions that execute the trajectory associated with the tool. From an agentic AI perspective, this approach separates learning into two levels: tool discovery is handled through unsupervised quantization of trajectories, while reasoning and decision-making are handled by the LLM. We test the approach in a partially observable dynamic 2D grid environment with an open vision-language model (Qwen3.6-35B-A3B). Pairing the geometry-derived tool library with an agent-centered zoom tool and a collision detection tool lets a fast, non-reasoning configuration match the goal-reaching rate of a much more costly chain-of-thought version, while cutting the cost of a decision from minutes to seconds.
\end{abstract}



\section{Introduction}
\label{sec:intro}

Large Language Models (LLMs) have become the de-facto controller in many recent agentic systems, chaining natural-language reasoning steps with external tool calls to solve tasks that go beyond pure text generation~\citep{yao_react_2023,schick_toolformer_2023,wei_chain-of-thought_2022}. Spatial reasoning, however, remains a recognized weak spot: LLMs are trained on token sequences and have no built-in notion of metric space, obstacles or continuous motion so they tend to fall back on superficial statistical regularities of text rather than genuine geometric reasoning about the environment they are asked to act in. This is problematic for any application that requires an agent to move through, or reason about, physical or simulated space - navigation, warehouse robotics, game-playing agents, and embodied assistants among them. Difficulties accumulate when the simulators used (for instance in a reinforcement learning framework) face uncertainties or use approximations about spatial movements or future scenarios.

Interestingly, humans do not solve navigation problems by directly reasoning over raw coordinates either. When giving directions inside a building, we do not say ``move 1.5 meters north, then 3.2 meters east''; we say \emph{``go straight ahead, take the elevator, go around the corner office, then turn left.''} The instruction is a short composition of a handful of reusable, generic strategies - go straight, take the elevator, turn, avoid an obstacle - each of which packages several nontrivial low-level motions into a single high-level decision. A closely related approach appears in recreational mathematics: solving a Rubik's cube does not require deriving new moves from scratch for every configuration. Instead, a small fixed library of move sequences, called \emph{algorithms} in the cubing community (e.g., \texttt{R\,U\,R'\,U\,R\,U2\,R'} or \texttt{F\,R\,U\,R'\,U'\,F'}, see \url{https://alg.cubing.net}), is reused across arbitrary starting positions; the solver's actual skill lies in recognizing \emph{which} algorithm applies to the current configuration, not in re-deriving the moves themselves for each starting configuration. We take this as our guiding analogy: if LLMs are weak at direct, low-level geometric planning, perhaps they should not be asked to do it. Instead, they can be asked to do what they are comparatively good at: recognizing patterns and selecting among a small set of well described, precomputed options.

This paper explores exactly this division of labor for spatial tasks. We equip an LLM-driven agent with a library of \emph{tools}, each one a short, reusable motion pattern extracted directly from the geometry of the environment rather than hand-designed or learned through reinforcement learning; we view these geometry-grounded modules as structured inductive biases. Concretely, we (i) sample many \emph{geodesic} trajectories - shortest, obstacle-aware paths - inside a partially observable 2D grid world, capturing the environment's topology and spatial structure, (ii) compress this large trajectory set into a small number of representative prototypes using a \emph{\textbf{vector quantization}} (VQ) procedure, each prototype becoming a reusable strategy, (iii) in an \textbf{LLM interpretation} step, use an LLM \emph{offline} to translate each prototype into a natural-language description of its geometric behavior, its strengths, and its risks, turning it into a callable \textbf{tool}, and (iv) use the same (or another) LLM \emph{online}, at decision time, for \textbf{LLM orchestration}: picking the tool best suited to the current partial observation and goal. Low-level actuation is then simply ``play out the chosen tool's fixed action sequence,'' mirroring the way a Rubik's cube solver executes a memorized algorithm once it has recognized the pattern.

From an agentic AI perspective, this design yields an \textbf{agentic separation} of two steps that are often entangled: \emph{skill discovery}, handled by unsupervised geometric quantization and requiring no labels or reward signal, and \emph{skill selection / reasoning}, handled by the LLM using its native strengths in language-conditioned decision-making. Our \textbf{key insight} is that LLMs appear to be more effective as \emph{orchestrators} of geometry-derived skills than as direct spatial planners.

We present the synthetic 2D grid world with moving obstacles and partial observability in Section~\ref{sec:env} and the methodology in Section~\ref{sec:method}; results
in Section~\ref{sec:results} 
show that routing decisions through the geometry-grounded tool library improves significantly the target-reaching success rate
over querying the LLM directly.

\section{Related work}
\label{sec:related}

Our approach sits at the intersection of three lines of work that are, to our knowledge, rarely combined in this particular way.

\textbf{LLM agents and tool use.} LLM-driven agents increasingly interleave natural-language reasoning with calls to external tools or actions, as in ReAct~\citep{yao_react_2023} and Toolformer~\citep{schick_toolformer_2023}, and benefit from decomposing a task into intermediate reasoning steps, as in Chain-of-Thought paradigm of~\citep{wei_chain-of-thought_2022}. These works treat tools mostly as external APIs (calculators, search engines, code interpreters); we instead \emph{derive} the tool set itself from the geometry of the environment, so that the LLM orchestrates a set of skills that is neither given a priori not invented by the LLM alone.

\textbf{Quantization and skill discovery.} Vector quantization has a long history as a way to compress a large or continuous dataset into a small number of representative prototypes, from classical VQ codebooks to VQ-VAE-style discrete representation learning~\citep{vandenoord_neural_2017} and to measure-theoretic versions~\citep{turinici_huber-energy_2024}  which can handle signed measures - a property that could be exploited to let new skills emerge from novel trajectory patterns rather than fixing the tool set in advance.

\textbf{Spatial reasoning and geodesics.} Geodesics - minimal paths that reflect the intrinsic (possibly curved, possibly obstacle-constrained) geometry of a space, as opposed to paths that are straight only in some extrinsic embedding - are a classical tool in geometry and robotics for capturing the true topology of an environment. A growing body of work documents that LLMs and vision-language models struggle with exactly this kind of geometric reasoning when it is left implicit in raw or visual input: \citet{patel_pavlick_2022} find that only the largest LMs learn to ground directional/spatial concepts in a grid world, and only from explicit examples; \citet{aghzal_ppnl_2023} show with their PPNL benchmark that GPT-4 plans spatially only when prompted to interleave reasoning and acting, and still fails at longer-horizon temporal reasoning; \citet{martorell_text2space_2025} show that grid-navigation success depends heavily on how coordinates are textually encoded, scaling with model size; \citet{sandfuchs_llms_2026} find that under strict partial observability only reasoning-tuned LLMs reliably reach the goal in text gridworlds, and still less efficiently than an oracle path; \citet{gao_cubebench_2025} use a Rubik's-cube benchmark to isolate long-horizon spatial state tracking under partial observation, reporting a uniform $0\%$ pass rate for current LLM agents once external solver tools are removed; and \citet{rodriguez_mazebench_2026} show that vision-enabled models solving maze images are largely performing brute-force path enumeration in token space rather than genuine visual planning. Complementary work instead sidesteps this weakness by keeping the LLM out of low-level geometric computation: \citet{zhu_ghost_2023} plan long-horizon Minecraft tasks over structured text actions and memory rather than raw pixels; \citet{song_guidellm_2024} and \citet{rahimi_visionbased_2025} each pair the LLM with a dedicated non-LLM module - a text-based topological map, or a vision-based localizer - that handles metric geometry so the LLM only reasons over a simplified representation; \citet{dang_wayfinding_2024} equip an LLM wayfinding agent with externally maintained spatial-memory representations instead of inferred geometry; and \citet{zhao_cityeqa_2025} tackle city-scale embodied question answering via a hierarchical planner/manager/actor agent built around an explicit object-centric cognitive map rather than a single LLM call over raw city-scale geometry. Rather than asking an LLM to reason directly about such geometric objects, which is precisely the kind of task token-based models handle poorly, we use geodesics purely as an offline data-generation and skill-discovery mechanism, and reserve the LLM's role for the language-native tasks of describing and selecting among the resulting skills.
\section{Grid-world environment}
\label{sec:env}

\begin{figure}[h]
\centering
\includegraphics[width=0.15\linewidth]{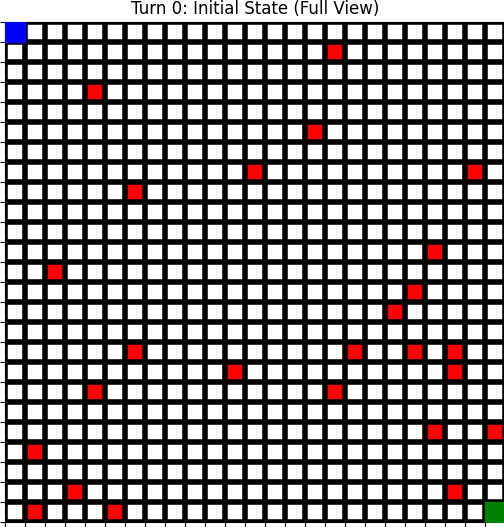}
\includegraphics[width=0.15\linewidth]{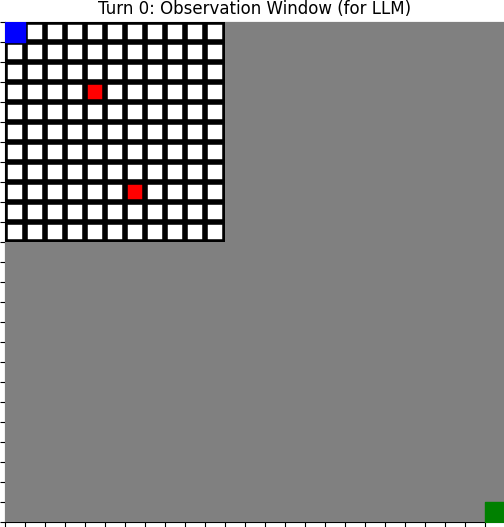}
\includegraphics[width=0.07\linewidth]{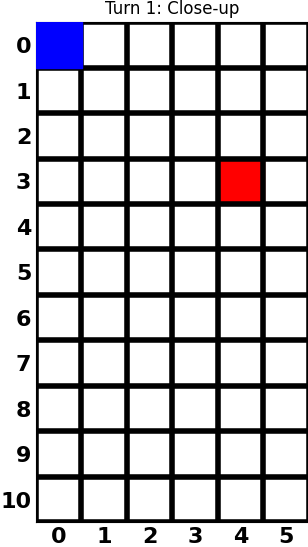}
 \hspace{0.6cm}
\includegraphics[width=0.15\linewidth]{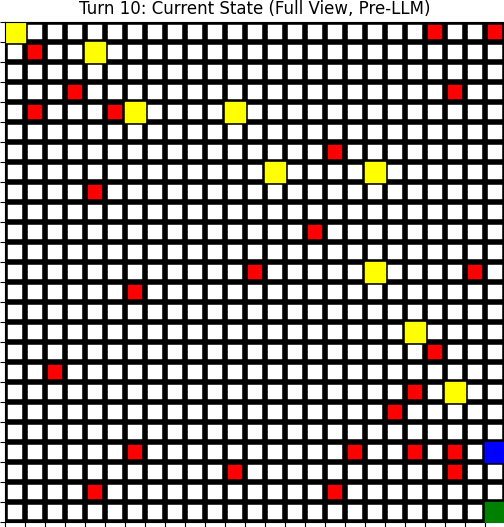}
\includegraphics[width=0.15\linewidth]{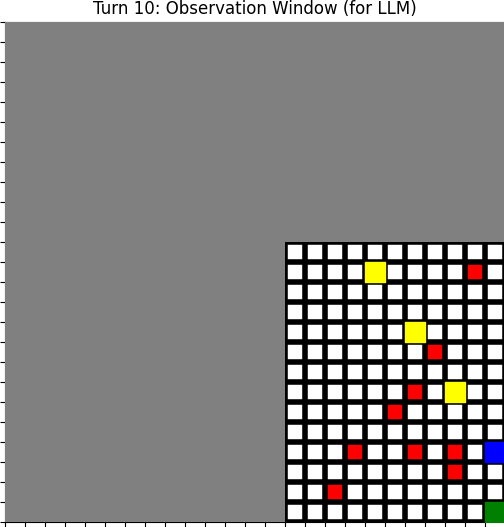}
\includegraphics[width=0.07\linewidth]{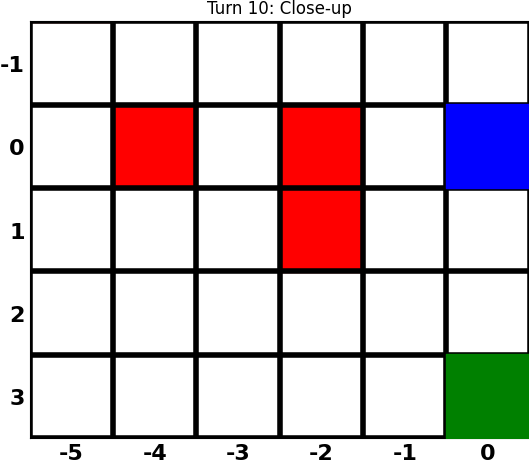}

\includegraphics[width=0.15\linewidth]{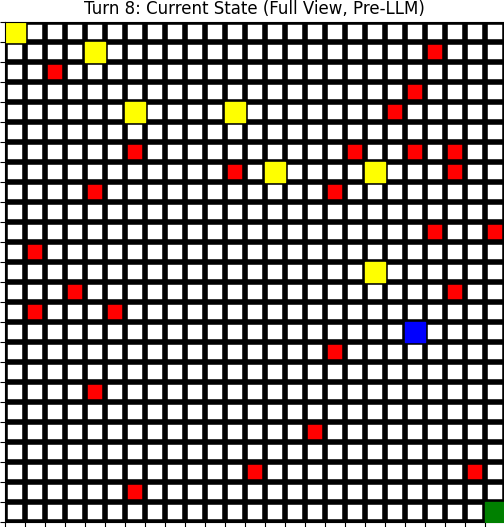}
\includegraphics[width=0.15\linewidth]{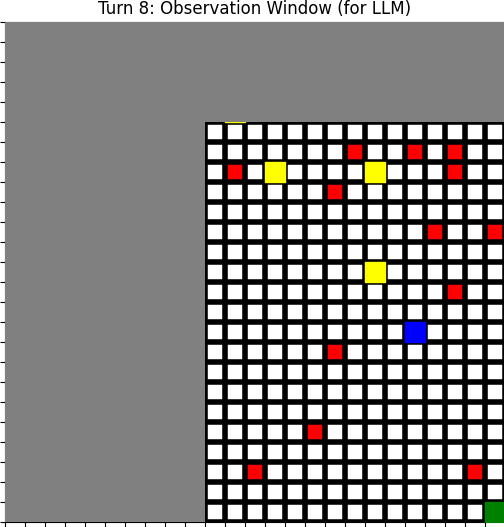}
\includegraphics[width=0.07\linewidth]{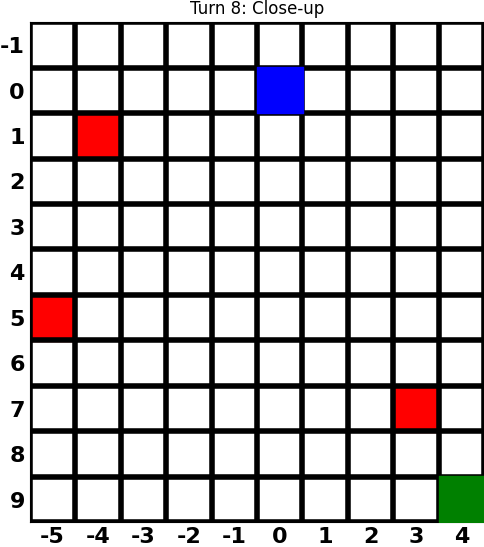}
\caption{All panels use the LLM Qwen3.6-35B-A3B. Top row, left to right: full grid at $t=0$, the partial observation given to the LLM at $t=0$, and its agent-centered close-up; then the full grid, partial observation and close-up at the last decision point, just before the target is reached (turn~10). Bottom row: an intermediate decision (turn~8) -- full grid, partial observation, and close-up. The agent only ever sees the observation window and the close-up; the full-grid panels are shown only to indicate what it could not see.}
\label{fig:env}
\end{figure}

To study this approach in a controlled setting we use a 2D grid environment loosely inspired by lane-crossing games such as Frogger. The environment illustrated in Figure~\ref{fig:env} is defined as follows:
\begin{itemize}
    \item \textbf{Agent:}
    \tikz{\fill[blue] (0,0) rectangle (1.5ex,1.5ex);}
    a single controlled cell (blue).
    \item \textbf{Cars:}
    \tikz{\fill[red] (0,0) rectangle (1.5ex,1.5ex);}
    dynamic obstacles that move upward by one cell per time step, representing traffic the agent must avoid.

    \item \textbf{Free space:}
    \tikz{\fill[white] (0,0) rectangle (1.5ex,1.5ex);\draw (0,0) rectangle (1.5ex,1.5ex);}
    cells that can be traversed safely at the current time.

    \item \textbf{Grey cells:}
    \tikz{\fill[gray] (0,0) rectangle (1.5ex,1.5ex);}
    unobserved cells, which may contain additional cars outside the agent's observation window.

    \item \textbf{Target:}
    \tikz{\fill[green] (0,0) rectangle (1.5ex,1.5ex);}
    the goal cell the agent must reach.

    \item \textbf{Yellow highlight:}
    \tikz{\fill[yellow] (0,0) rectangle (1.5ex,1.5ex);}
    marks the time steps at which a  decision was taken.

   \item \textbf{Actions:} wait, up, down, left, right - five primitive actions available at every time step.

    \item \textbf{Partial observability:} only a local window centered on the agent
    (\tikz{\fill[blue] (0,0) rectangle (1.5ex,1.5ex);})
    is visible; cells outside this window are grey and unknown.

\end{itemize}

The {
\textbf{objective}} is to 
reach the goal cell, if possible with minimum expected number of movements and elapsed time but imperatively avoiding collisions with cars.

To this end we focus on  {
\textbf{multi-step decisions}};
we cannot take one decision at a time because the environment evolves quickly relative to the decision frequency of the LLM  - the same constraint that forces landers and deep-space rovers to commit to sequences of actions rather than await step-by-step commands from Earth. To keep the number of costly (in terms of money or wall-clock time) LLM queries manageable, at each decision point the agent commits to and executes a fixed sequence of five future actions regardless of how the environment evolves in between; only then the next decision is taken. This is precisely the setting that motivates our tool design: each tool introduced below is such a five-action sequence, pre-computed offline so that, online, the LLM only needs to pick one rather than plan it step by step.

\section{Method}
\label{sec:method}

Our architecture is built in several stages, executed in the order below: geodesic collection, offline quantization, tool creation, and online orchestration.

\subsection{Geodesics as shortest-path connections}
\label{sec:geodesics}

\begin{figure}
    \centering
    \includegraphics[height=3cm]{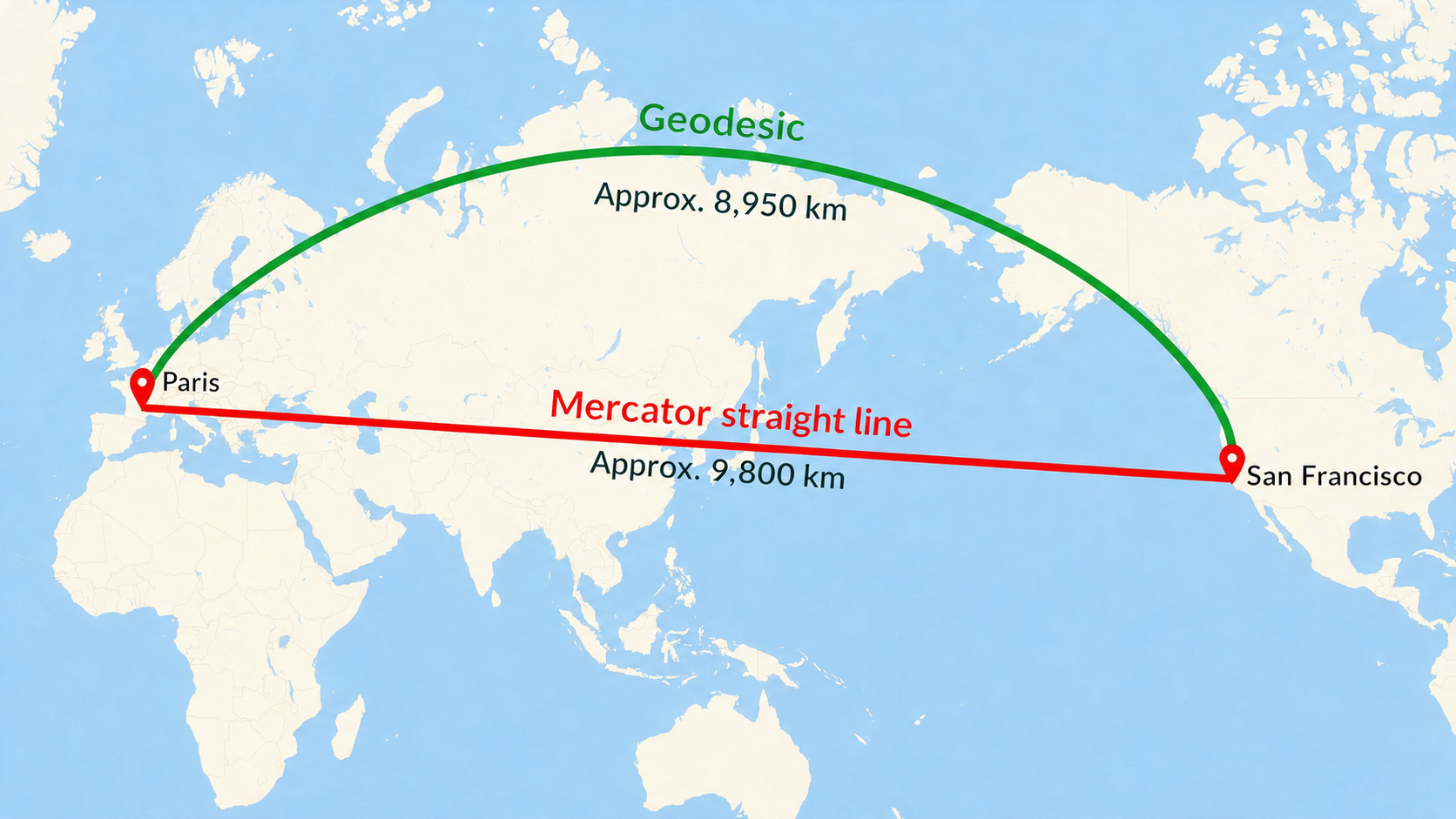}
\hspace{1cm}
    \includegraphics[height=1cm]{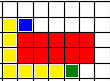}
    \caption{
    {\bf Left:} geodesic between two cities on the 3D Earth surface: 
    great-circle geodesic in green versus the naive straight line on a 2D Mercator projection in red: the shortest path in the underlying geometry is not the visually straight one.
    {\bf Right:} a geodesic in our environment that has to dodge  cars (in red), assuming for graphical purposes that cars are static (do not move) : initial cell is blue, target green and geodesic in yellow. More direct routes are not available.
    }
    \label{fig:geodesic}
\end{figure}

A geodesic captures the intrinsic geometry of a space by defining the minimal path between two locations; curvature and obstacles influence which path is actually shortest. On a flat, obstacle-free plane a geodesic is just a straight line, but as soon as the space is curved or obstacles are present, the shortest path can look very different from the naive straight-line, see examples in 
Figure~\ref{fig:geodesic}.
In our grid world, we treat the agent's start cell, the goal cell, and the (possibly moving) obstacles as defining a discrete metric space, and compute geodesics as shortest admissible paths in this space - i.e., paths that reach the goal in as few steps as possible while never crossing a car. Because obstacles move and the environment is only partially observed, many such geodesics are sampled under different obstacle realizations and different partial-information conditions, so that the resulting set reflects the range of spatial strategies that are useful across plausible configurations of the environment, rather than a single trajectory tailored to one specific instance.

\subsection{Vector quantization of trajectories}
\label{sec:vq}
Sampling many geodesics is useful for capturing the environment's topology, but it also produces far more trajectories than can be handed, individually, to an LLM as candidate actions: the resulting library would be too large to describe or to choose from efficiently. We therefore compress the sampled trajectories using \emph{vector quantization} (VQ): given a large set of trajectories, VQ selects a small number of representative prototypes that best summarize the whole set, in the same way that classical VQ or VQ-VAE codebooks summarize a high-dimensional dataset with a handful of codewords~\citep{vandenoord_neural_2017}.

\begin{figure}[h]
\centering
\includegraphics[width=0.2\linewidth]{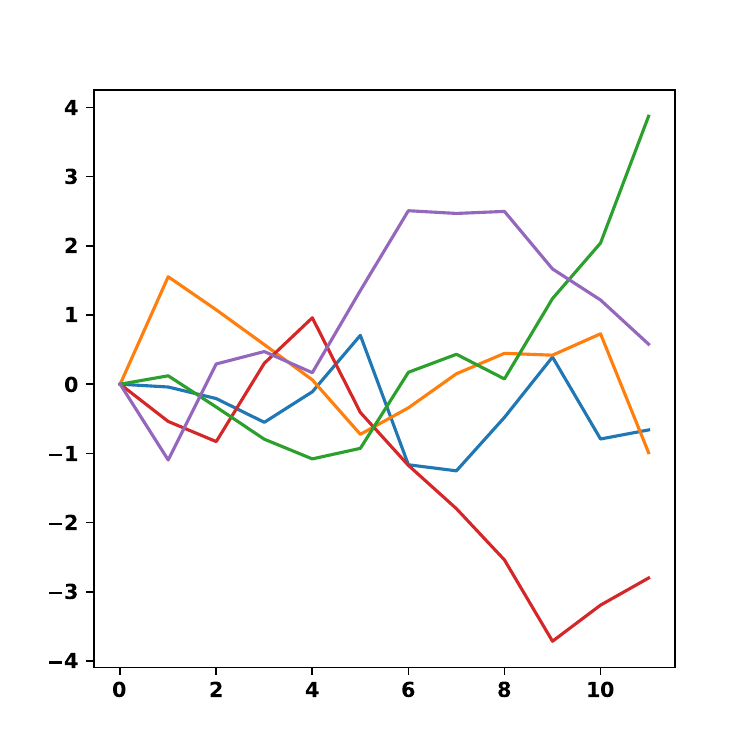}
\includegraphics[width=0.2\linewidth]{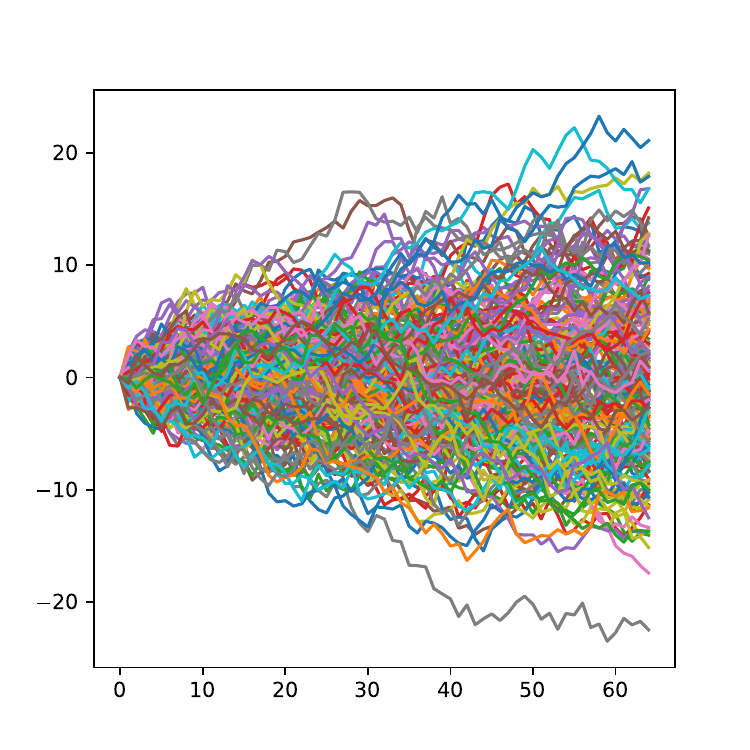}
\caption{Quantization of standard Brownian motion (also called cubature), used here as a sanity-check illustration of the VQ machinery 
on a classical stochastic-process benchmark 
before applying it to grid geodesics. 
 Left: $5$ representative trajectories selected out of $11$-step paths of a discrete Brownian motion. Right: $200$ 
representative 
trajectories 
selected out of $64$-step Brownian motions, illustrating that the same procedure scales to larger, denser trajectory sets. \citep{turinici_huber-energy_2024,turinici_diversity_2024}.
of a Brownian motion.
}
\label{fig:brownian}
\end{figure}

Concretely, we treat the empirical distribution of sampled geodesics as a (possibly signed, cf. ~\citep{turinici_huber-energy_2024,turinici_diversity_2024}) distribution and apply a 
measure quantization procedure (like K-medoids) which finds $K$ representative trajectories in the given set.
Figure~\ref{fig:brownian} illustrates the general idea on a simpler, well-understood test case - quantization of standard Brownian motion paths, also known as cubature.

\begin{figure}[h]
\centering
\includegraphics[width=0.3\linewidth]{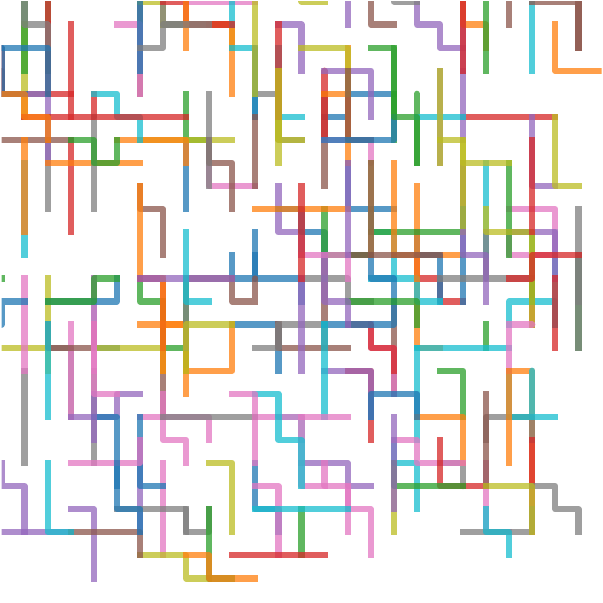}\hfill
\raisebox{0.04\linewidth}{%
\includegraphics[width=0.125\linewidth]{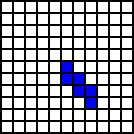}%
\includegraphics[width=0.125\linewidth]{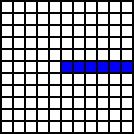}%
\includegraphics[width=0.125\linewidth]{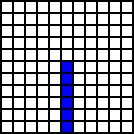}%
\includegraphics[width=0.125\linewidth]{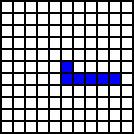}%
\includegraphics[width=0.125\linewidth]{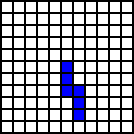}}
\caption{{\bf Left: } the $250$ sampled geodesic trajectories in the grid environment, overlaid; each color is one geodesic. Trajectories are drawn here from their true start cells; in the pipeline all start positions are reset to the origin before quantization.
{\bf Right: } their vector quantization into $K=5$ representative five-step trajectories, one prototype per panel (\texttt{DRDRD}, \texttt{RRRRR}, \texttt{DDDDD}, \texttt{DRRRR}, \texttt{DDRDD} -- named by their action letters U/D/L/R/W), each of which becomes a reusable tool. To this library we add the five primitive tools \texttt{UWWWW}, \texttt{DWWWW}, \texttt{LWWWW}, \texttt{RWWWW}, \texttt{WWWWW} (one primitive move then four waits). These tools are then used for the results in Figure~\ref{fig:results}.
}
\label{fig:gridvq}
\end{figure}

Applied to the grid environment, we sample $250$ geodesic trajectories under varying obstacle configurations and quantize them down to $K=5$ representative prototypes (Figure~\ref{fig:gridvq}).

Each of the $K$ prototypes gives a candidate tool, described in natural language in Section~\ref{sec:tools} and made available to the online orchestrator in Section~\ref{sec:orchestration}.
To this library we also add the primitive actions (a primitive action followed by four ``wait'' actions) in case the target is already close or a precise move is needed.

\begin{remark}
A feature of this construction, which we record here for future reference but do not evaluate in the experiments below (where the tool set is fixed), is that it need not fix the tool set once and for all. Because the quantization can be formulated over \emph{signed} measures rather than only probability measures, cf.
\citep{turinici_huber-energy_2024}, 
one could monitor, at any point during data collection, the discrepancy between the empirical measure of newly observed geodesics and the signed measure induced by the currently selected prototypes. When this discrepancy is large - i.e., when new trajectories are not well represented by any existing prototype - a new representative trajectory would be added to the tool set. In this way new skills would emerge precisely when the agent's behavior contrasts with previously captured patterns, rather than requiring the full tool library to be enumerated in advance.
\end{remark}

\subsection{From quantized trajectories to LLM tools}
\label{sec:tools}

Each of the $K$ quantized five-step action sequences is, on its own, meaningless to an LLM: it is just a list of moves such as \texttt{down, right, down, right, down}. To make it usable as a decision-time tool, we run an LLM \emph{offline}, once per prototype, and prompt it to describe the trajectory in natural language: what net displacement it produces, what tactical advantages and risks it carries with respect to the moving-car dynamics of the environment, and under what circumstances it should or should not be selected. This offline pass converts a purely geometric object into a self-contained \textbf{TOOL}; note that in principle this callable unit can, at run time, also report spatial metrics such as the remaining distance to the goal and locally observed obstacles, in addition to the natural-language description attached to it.
We report in appendix one such tool description. When the deterministic collision check of Section~\ref{sec:results} is used, this offline description is shortened to just the net displacement and a one-line strategic summary, since the collision and timing reasoning is then handled in code rather than by the LLM.

Each of the $K$ tools is described in the same way, so that the online orchestrator always has access to a small library of $K$ short, natural-language tool descriptions, rather than raw trajectories. As discussed above, we also add the primitive actions to this library.

\subsection{Online orchestration}
\label{sec:orchestration}

At decision time, the LLM does not plan a path from scratch. Instead, it is given 
\begin{itemize}
    \item 
the current partial observation i.e., the local window around the agent, including visible cars and, if within range, the target; in practice we use a vision-enabled LLM, Qwen3.6-35B-A3B, and input both the partial observation and an agent-centered close-up (``zoom'') of it, cf. Figure~\ref{fig:env}. The close-up (second input image) was required for Qwen3.6-35B-A3B to produce good results.
\item  the description of the goal (i.e., its color),
\item the natural-language tool descriptions produced offline in Section~\ref{sec:tools};
\item optionally, a collision detection tool that removes tools that would collide with observed cars (but not with unobserved ones).
\end{itemize}
Its task is purely one of \emph{selection}: pick the single tool whose described behavior best matches the current situation, e.g., preferring a down-right staircase tool when the lower-right is clear and the target lies down-right, or the all-wait tool when no safe move is currently available. Once a tool is selected, its fixed action sequence is executed open-loop by the low-level controller, irrespective of how the environment evolves during those five steps, and the cycle repeats at the next decision time.

This separation is the central agentic design choice of the paper: geometric skill discovery (Sections~\ref{sec:geodesics}-\ref{sec:tools}) is unsupervised and requires no LLM involvement at all, while the LLM is only ever asked to do two language-native tasks, describe a trajectory, and choose among descriptions, rather than to reason directly about coordinates, distances, or collision geometry.

\section{Results}
\label{sec:results}

We evaluate the full pipeline (geodesic sampling $\to$ VQ into $K$ tools $\to$ offline LLM tool descriptions $\to$ online LLM orchestration) against a baseline in which the same LLM is queried directly on the raw grid observation, with no geodesic tools, no quantization, and no natural-language skill descriptions i.e., the LLM must propose the next five actions itself from the same visual/textual input. Implementation is provided in the github repository 
\url{https://github.com/gabriel-turinici/quantized_geodesics_agents}.

\begin{figure}
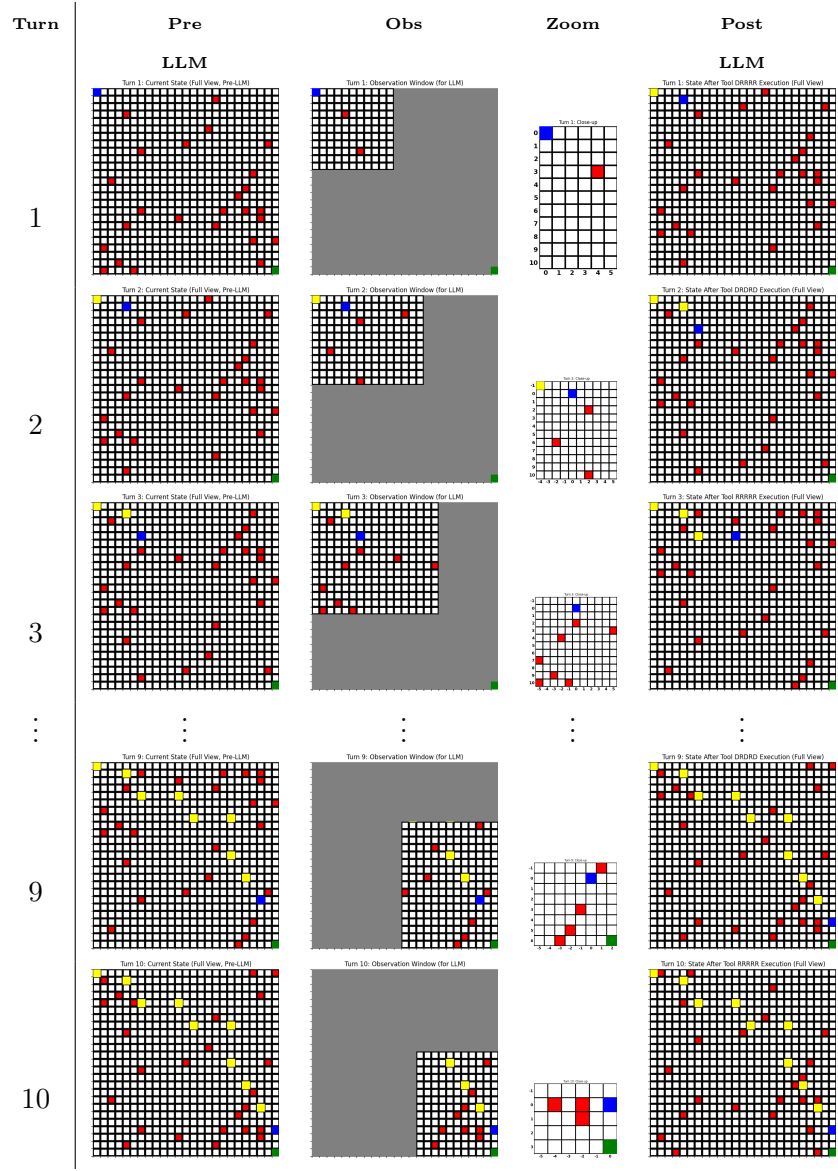

\centering
\begin{tabular}{c|cccc}
\textbf{\tiny Turn} &
\textbf{\tiny Pre} &
\textbf{\tiny Obs} &
\textbf{\tiny Zoom} &
\textbf{\tiny Post}
\\
\ & \textbf{\tiny LLM} &
\textbf{\tiny \ } &
\textbf{\tiny \ } &
\textbf{\tiny LLM} \\

\turnrow{1}
\turnrow{2}
\turnrow{3}

\vdots & \vdots & \vdots & \vdots & \vdots \\

\turnrow{9}
\turnrow{10}
\end{tabular}
\caption{Sequence of decision turns: full grid before the decision, the partial observation given to the LLM, its agent-centered close-up (``zoom''), and the full grid after executing the selected tool's five-action sequence, until the target is reached at turn~10. The LLM is Qwen3.6-35B-A3B.}
\label{fig:results}
\end{figure}

Figure~\ref{fig:results} shows a representative episode (LLM: Qwen3.6-35B-A3B): for a sequence of decision points ("turns"), it displays the full grid state just before the LLM decision (Pre-LLM), the partial observation actually given to the LLM (Obs) together with its agent-centered close-up (Zoom), and the full grid state after the selected tool's five-action sequence has been executed (Post-LLM). The agent (blue cell) is only ever shown its local observation window and the close-up of it; the full-grid panels are shown here purely to illustrate what the agent could not see.

We tested several LLMs but most failed to ever reach the target, so we focus on the vision-enabled model {\bf Qwen3.6-35B-A3B}, used both for offline tool description and for online tool selection in all runs. A second, larger frontier vision model was also tried on a small number of episodes and showed the same qualitative trends - tools helping over direct querying - but our compute budget did not allow a full evaluation, so we report quantitative results for Qwen3.6-35B-A3B only. The results (Table~\ref{tab:success_rate}) can be summarized in three observations:
\begin{itemize}
    \item 
 first two lines: when thinking mode is disabled and the collision tool absent the model fails. The detailed logs (not shown here but reproducible with the github code) show that the model tries to reproduce the collision detection computations manually, reasoning one step at a time but gets lost because of the complexity of the task
\item
 lines 3 and 4: when thinking mode is enabled: the absence of the collision-detection tool is compensated by a high effort of around 20k tokens per turn and takes a non-negligible amount of time; 
the geodesic-VQ-tool pipeline reaches the target successfully
 more often than the same LLM queried directly on the raw grid observation under matched data and query budget; 
\item 
 lines 5 and 6 :  once a collision validation tool filters out unsafe tools before the LLM decides (see also Appendix~\ref{app:repro}), results come back to optimal and a fast non-thinking configuration matches the success rate of full thinking mode at a small fraction of the cost. 
\end{itemize}

\begin{table}[htbp!]
\centering
\small
\setlength{\tabcolsep}{5pt}
\begin{tabular}{c|c|c|c|c|c|c}
\hline \hline \noalign{\vskip 0.15cm}
\multicolumn{7}{c}{\textbf{Success rate for Qwen3.6-35B-A3B}} \\
 \noalign{\vskip 0.15cm}
\hline \hline \noalign{\vskip 0.15cm}
           &               & Collision  &  Zoom        & Average     & Average   &             \\
  Thinking &   Geodesic    &  check     &  (close-up)  & time        & tokens    & Success     \\
           &   tools       &  tool      &  tool        & per         & generated & $\%$, ($N$) \\
           &               &            &              & turn        & per turn  &             \\
\noalign{\vskip 0.15cm}
\hline \hline
\noalign{\vskip 0.35cm}
\no   & \no  & \no  & \yes & $\sim 20.5$ sec.      &  1200     & $2\%\ (100)$ \\
\no   & \yes & \no  & \yes & $\sim 17$ sec.     & 850      & $2\%\ (100)$ \\
\noalign{\vskip 0.35cm}
\yes  & \no  & \no  & \yes & 5\,min.           & 19k   & $54\%\ (100)$ \\
\yes  & \yes & \no  & \yes & 5.5\,min.         & 20k   & $66\%\ (100)$ \\
\noalign{\vskip 0.35cm}
\yes  & \yes & \yes & \yes & $\sim 9.5$ sec.     & 450   & $65\%\ (100)$ \\
\no   & \yes & \yes & \yes & $\sim 5$ sec.   & 200   & {\bf 68\%\ (100)}\\
\noalign{\vskip 0.15cm} 
\hline \hline
\end{tabular}
\caption{Grid-navigation success rate for Qwen3.6-35B-A3B across reasoning
mode and the geometry-derived components: the geodesic tool library, a code
collision check that filters unsafe tools before the LLM decides, and the
agent-centered close-up image. ``Time / turn'' is the wall-clock cost of one
decision. All
rows use the close-up image, which Qwen3.6-35B-A3B requires, without it
results are very poor. The final row shows that, with the collision check
and the close-up in place, the geodesic tools reach top quality (comparable
to full thinking mode) at a small fraction of the time.}
\label{tab:success_rate}
\end{table}

This improvement in target-reaching success supports our central hypothesis: the bottleneck for LLM spatial reasoning is not necessarily the LLM's decision-making \emph{per se}, but the representation on which that decision is made. When the LLM is asked to reason directly over a raw grid of colored cells, it must implicitly perform geometric operations (shortest paths, alternatives, collision timing) that are known to be a poor match for token-based reasoning. When the same decision is reduced to choosing among a handful of pre-computed, geometrically grounded, natural-language-described options, the LLM's comparative strength - language-conditioned selection under a stated objective - is what actually gets exercised, and performance improves markedly under matched conditions. A further consequence is on cost: once a deterministic collision check removes unsafe tools before the query, the remaining choice is easy enough that chain-of-thought reasoning is no longer needed, and a non-thinking model reaches the same success rate as full thinking mode while cutting the per-decision cost from minutes to seconds (Table~\ref{tab:success_rate}, last row).

\section{Discussion and limitations}
\label{sec:discussion}

The grid-world setting used here is intentionally simple: it isolates the core question of whether decoupling geometric skill discovery from LLM-based orchestration helps spatial reasoning, without the confound of high-dimensional perception or continuous control. Several simplifications should be kept in mind when extrapolating the result. First, the environment is fully discrete and low-dimensional (a 2D grid with a handful of moving obstacles), so the offline quantization problem is comparatively easy; scaling to continuous state spaces, 3D navigation, or robotic manipulation will require replacing the discrete geodesic search and quantization with continuous analogues, which is a natural direction for future work. Second, the tool library size ($K=5$) was chosen for tractability of natural-language description and orchestration, not tuned systematically; how success rate and description quality scale with $K$, and how the signed-measure mechanism performs at triggering new skills in a longer-running, non-stationary environment, are open questions. Third, our comparison to a ``direct-query'' baseline uses the same LLM in both conditions, which isolates the effect of the tool-based representation but does not yet test whether smaller, possibly non-vision LLMs would benefit similarly from the same architecture. Fourth, the offline LLM tool descriptions were generated once and used verbatim; whether iteratively refining descriptions based on downstream orchestration failures would further improve performance is left for future work.
 Finally, we report a single LLM (Qwen3.6-35B-A3B); a larger frontier model was spot-checked with consistent trends but not fully evaluated for budget reasons.

A further observation, based on data not shown here, is that a major failure mode of the online orchestrator is perception rather than planning: the vision-enabled LLM occasionally misreads the position of a nearby car by one cell, which is enough to turn an otherwise-safe tool choice into a collision. Even large vision-enabled models remain unreliable at this kind of precise, cell-level reading of the observation image, and prompt engineering alone did not remove these errors.

\section{Conclusion}
\label{sec:conclusion}

We presented a protocol that separates spatial skill discovery from spatial decision-making in LLM-driven agents. Geodesic trajectories sampled from a partially observable grid environment are compressed via vector quantization into a small set of representative, reusable tools; an LLM is used offline to translate each such strategy into a natural-language tool description, and online to select among these descriptions given the current state and goal. This pipeline improves the target-reaching success rate compared to querying the same LLM directly on raw observations and, when paired with a deterministic collision filter, matches the accuracy of an expensive thinking-mode configuration at a fraction of its wall-clock cost, suggesting that LLMs are more effective as orchestrators of geometry-derived skills than as direct spatial planners. We see this as an argument supporting a broader agentic-AI design principle - let unsupervised, domain-appropriate methods discover skills, and let the LLM do what it does best: describe and choose between them - and hope to test its generality on higher-dimensional, continuous, and robotic navigation and manipulation settings in future work.


\bibliographystyle{plainnat}
\bibliography{sample}\vspace{0.75in}

\clearpage \newpage 

\appendix
\section{Technical appendices and supplementary material}

\subsection{Example of tool description}

The box below shows the tool description, generated automatically. 
The LLM is Qwen3.6-35B-A3B; when the deterministic collision check is enabled the
offline prompt is shortened to just the net displacement and a one-line
strategic summary; prompts are in the repository.

\begin{verbatim}
  Each tool is a list of actions that affect the position of the agent. When
  a tool is selected all its actions are implemented until collision, 
  reaching the goal, or exhaustion of the list. Each tool is named by its 
  actions, one letter per step: U=up (-1,0), D=down (1,0), L=left (0,-1), 
  R=right (0,1), W=wait (0,0).

  Available tools are:
  
 ====================tool WWWWW ====================:
 total displacement: no movement
action list: hold five steps
strategy: does nothing 
 ====================end description tool WWWWW ====================:


 ====================tool UWWWW ====================:
 total displacement: one row up, no sideways change
action list: up once, then hold four steps
strategy: a single step up followed by waiting 
 ====================end description tool UWWWW ====================:


 ====================tool DWWWW ====================:
 total displacement: one row down, no sideways change
action list: down once, then hold four steps
strategy: a single step down followed by waiting 
 ====================end description tool DWWWW ====================:


 ====================tool LWWWW ====================:
 total displacement: one column to the left, no vertical change
action list: left once, then hold four steps
strategy: a single step left followed by waiting 
 ====================end description tool LWWWW ====================:


 ====================tool RWWWW ====================:
 total displacement: one column to the right, no vertical change
action list: right once, then hold four steps
strategy: a single step right followed by waiting 
 ====================end description tool RWWWW ====================:


 ====================tool DRDRD ====================:
 total displacement: three rows down and two columns to the right
action list: down once, then right once, then down once, then right once, 
then down once strategy: a staircase pattern moving steadily toward the 
bottom-right 
 ====================end description tool DRDRD ====================:


 ====================tool RRRRR ====================:
 total displacement: five columns to the right, no row change
action list: right five times
strategy: a straight run towards the right side 
 ====================end description tool RRRRR ====================:


 ====================tool DDDDD ====================:
 total displacement: five rows down, no sideways change
action list: down five times
strategy: a straight vertical run downwards 
 ====================end description tool DDDDD ====================:


 ====================tool DRRRR ====================:
 total displacement: one row down and four columns to the right
action list: down once, then right four times
strategy: a single step down followed by a straight run to the right 
 ====================end description tool DRRRR ====================:


 ====================tool DDRDD ====================:
 total displacement: four rows down and one column to the right
action list: down twice, then right once, then down twice
strategy: a vertical descent with a single rightward step 
 ====================end description tool DDRDD ====================:
\end{verbatim}

\subsection{Reproducibility details}
\label{app:repro}

All code, prompts, tool files, and per-run logs are in the repository
\url{https://github.com/gabriel-turinici/quantized_geodesics_agents};
we summarize here what is needed to read the paper.

\textbf{Environment.} A $25\times 25$ grid; the agent starts near the top-left
and the target is a fixed cell near the bottom-right. Cars move up by one row
per step and wrap from the top row back to the bottom row; the agent does not
wrap and grid edges are hard walls (a move into a wall is absorbed). A collision
is a car entering the agent's cell, or the agent and a car swapping cells, on
the same step. The agent observes only a local window (half-size $10$) centered
on it; everything else is unobserved. Each decision commits the next five
primitive actions, executed open-loop, before the next decision is taken.

\textbf{Tools.} $250$ geodesics are sampled under randomized car configurations
and partial information, reset to a common origin, and quantized with K-medoids
to $K=5$ prototypes (Figure~\ref{fig:gridvq}), named by their action letters
(U/D/L/R/W). The five one-move primitives (one move then four waits) are added,
giving a $10$-tool library.

\textbf{Collision-check tool.} When enabled, before the LLM is queried a
deterministic routine replays each candidate tool against only the cars
currently inside the observation window and discards any tool that would
collide. If no tool survives, the agent waits; if exactly one survives it is
taken with no LLM call; otherwise the surviving tools (and their descriptions)
are the only options offered to the LLM, whose answer is further constrained to
that set. This removes collision geometry from the LLM's task and is what allows
the fast non-thinking configuration in Table~\ref{tab:success_rate}. Note that there are situations when cars are not visible so that collision detection tool is not always guaranteed to work, it only gives the best conclusion given available data. 

\textbf{Close-up image.} The ``zoom'' is a fixed-size crop of the observation
centered on the agent, rendered at higher resolution; it is passed alongside the
wide observation whenever the observation is passed
(Figures~\ref{fig:env},~\ref{fig:results}).

\textbf{Evaluation.} An episode succeeds if the agent occupies the target cell
at any step, and fails on collision or after $30$ decisions. Each table cell
reports successes over episodes ($N$ given); episode counts differ across
conditions because of the compute budget. Both the offline tool descriptions and
the online selection use the same model, Qwen3.6-35B-A3B.

\section{More detailed trajectory plot}
\begin{figure*} \centering \setlength{\tabcolsep}{1pt} \begin{tabular}{ccccc} \textbf{\tiny Turn} & \textbf{\tiny Pre LLM} & \textbf{\tiny Obs} & \textbf{\tiny Zoom} & \textbf{\tiny Post LLM} \\
\tiny 1 & \includegraphics[width=0.26\linewidth]{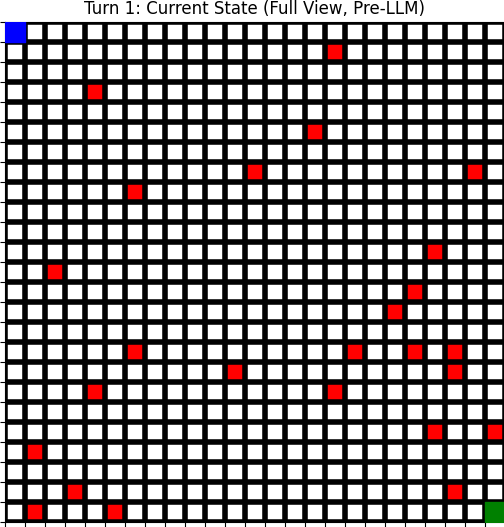} & \includegraphics[width=0.26\linewidth]{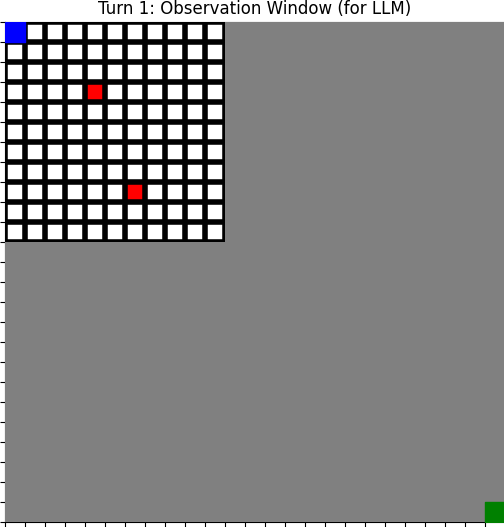} & \includegraphics[width=0.12\linewidth]{turn_1_zoom_view_for_llm.png} & \includegraphics[width=0.26\linewidth]{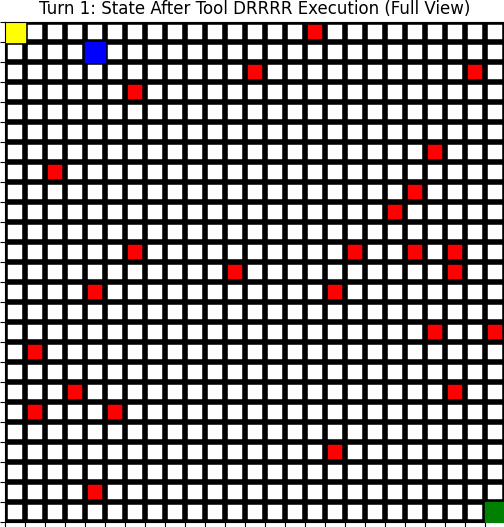} \\ \tiny 2 & \includegraphics[width=0.26\linewidth]{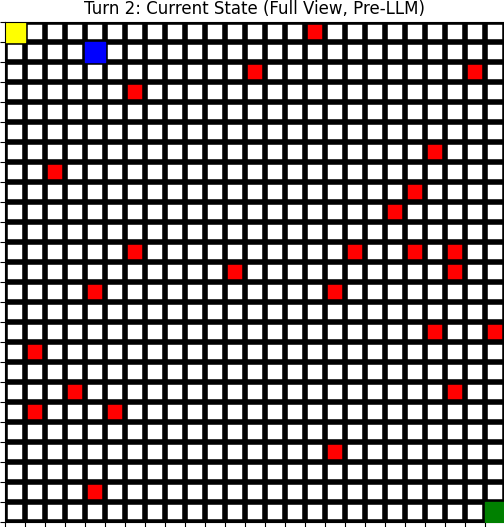} & \includegraphics[width=0.26\linewidth]{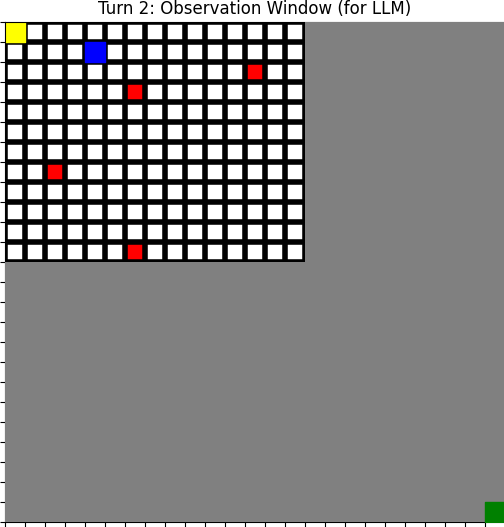} & \includegraphics[width=0.12\linewidth]{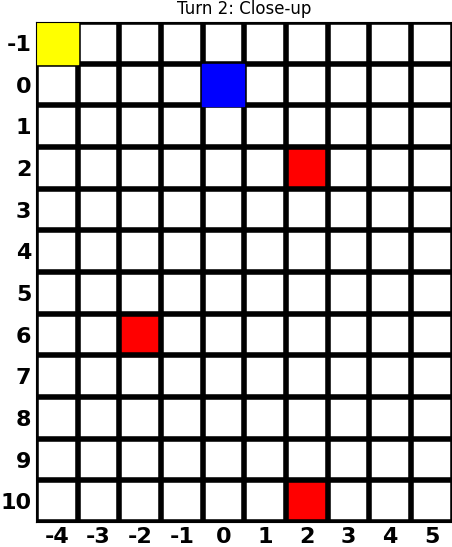} & \includegraphics[width=0.26\linewidth]{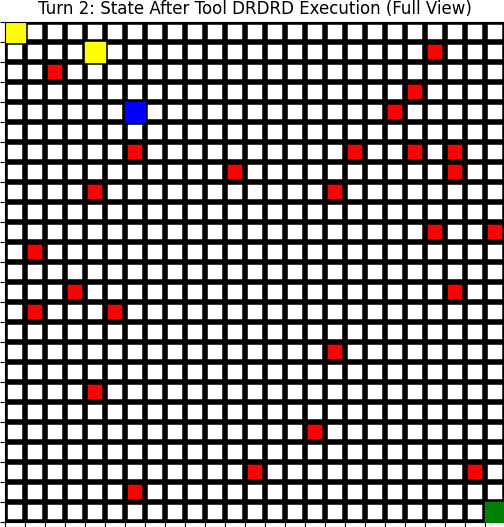} \\ \tiny 3 & \includegraphics[width=0.26\linewidth]{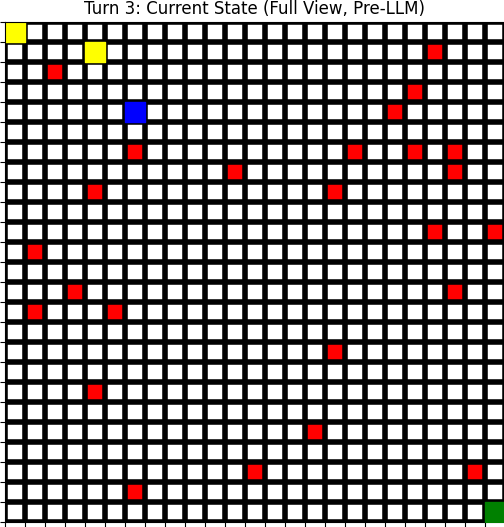} & \includegraphics[width=0.26\linewidth]{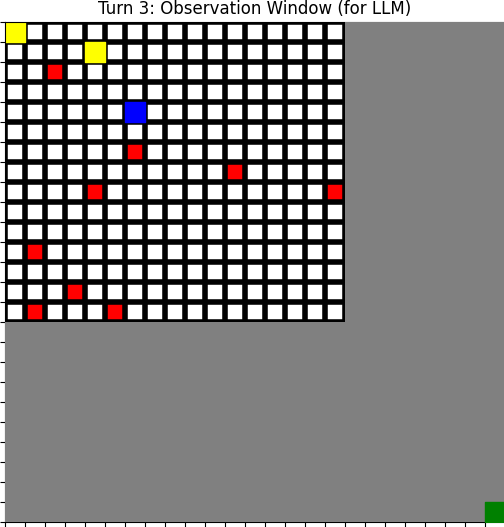} & \includegraphics[width=0.12\linewidth]{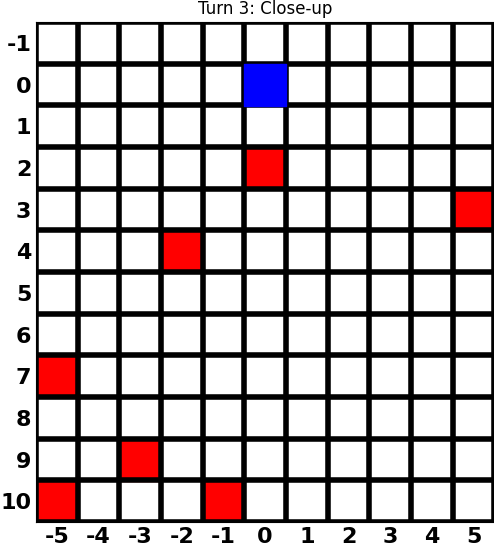} & \includegraphics[width=0.26\linewidth]{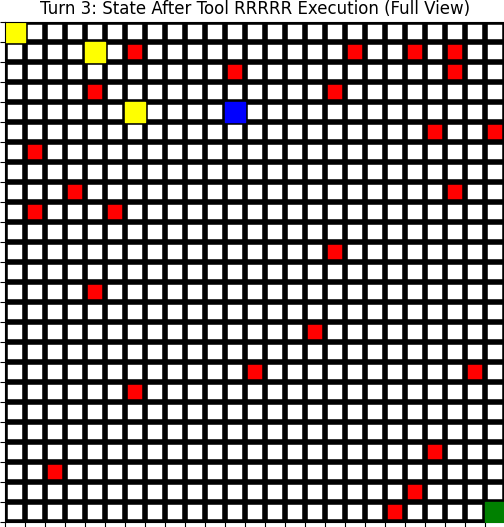} \\ \tiny 4 & \includegraphics[width=0.26\linewidth]{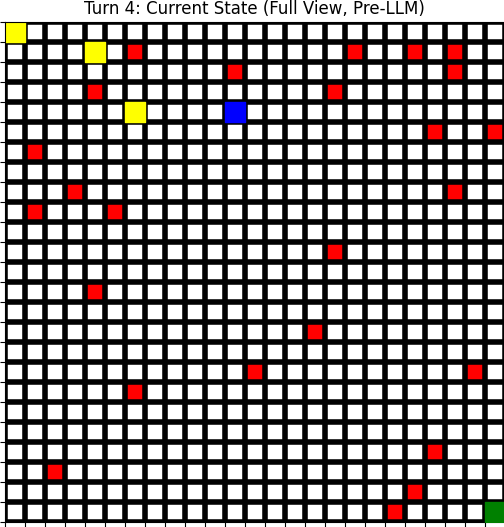} & \includegraphics[width=0.26\linewidth]{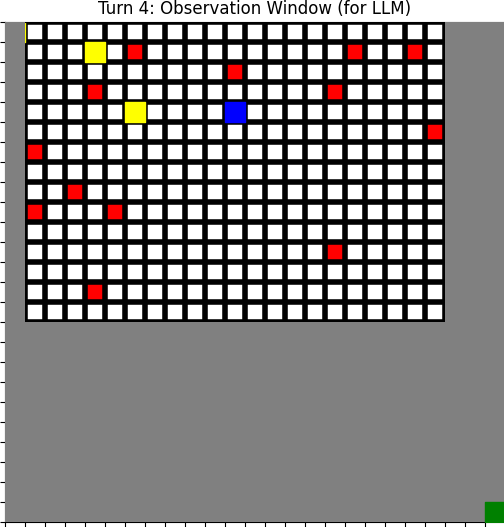} & \includegraphics[width=0.12\linewidth]{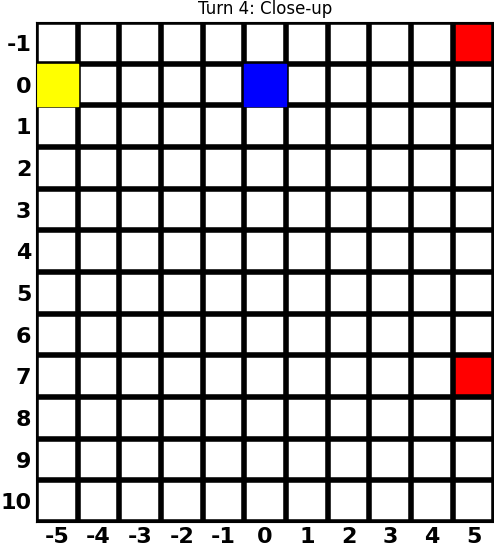} & \includegraphics[width=0.26\linewidth]{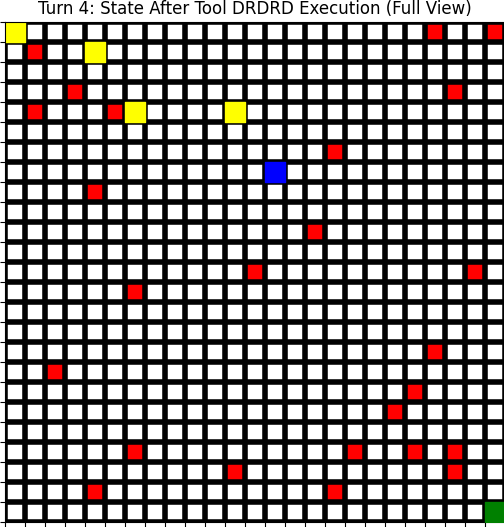} \\ \tiny 5 & \includegraphics[width=0.26\linewidth]{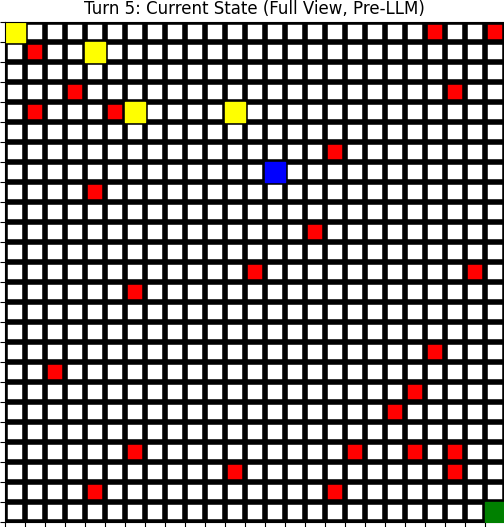} & \includegraphics[width=0.26\linewidth]{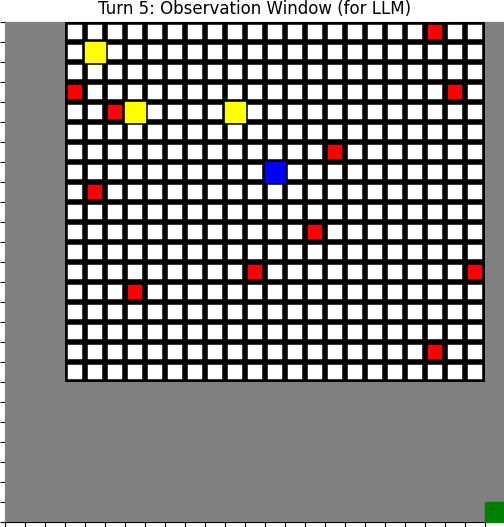} & \includegraphics[width=0.12\linewidth]{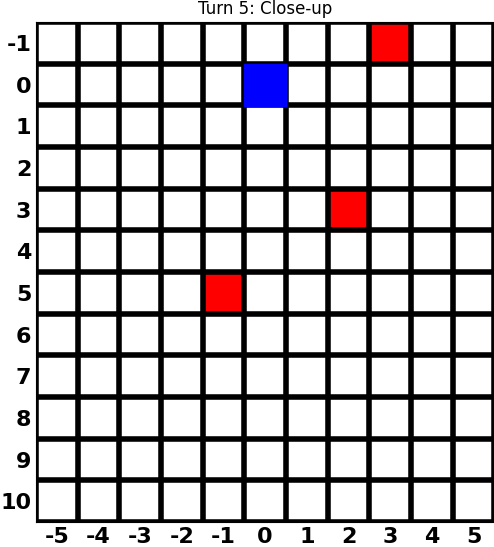} & \includegraphics[width=0.26\linewidth]{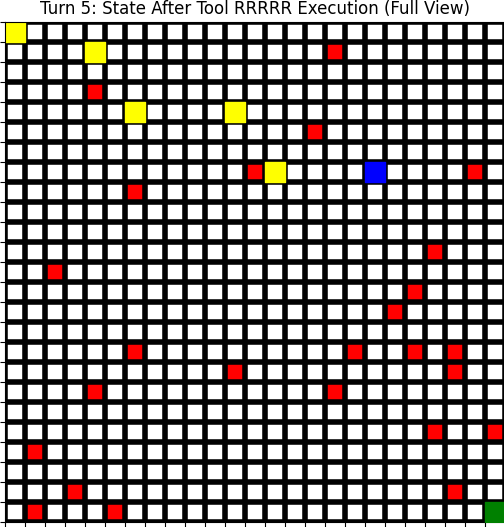}
\end{tabular} \caption{Detailed version of Figure~\ref{fig:results} (columns: full grid pre-decision, LLM observation, agent-centered close-up, full grid post-execution). Turns 1--5. LLM: Qwen3.6-35B-A3B.} \label{fig:results_detailed_portrait1to5} \end{figure*}

\begin{figure*} \centering \setlength{\tabcolsep}{1pt} \begin{tabular}{ccccc} \textbf{\tiny Turn} & \textbf{\tiny Pre LLM} & \textbf{\tiny Obs} & \textbf{\tiny Zoom} & \textbf{\tiny Post LLM} \\
\tiny 6 & \includegraphics[width=0.26\linewidth]{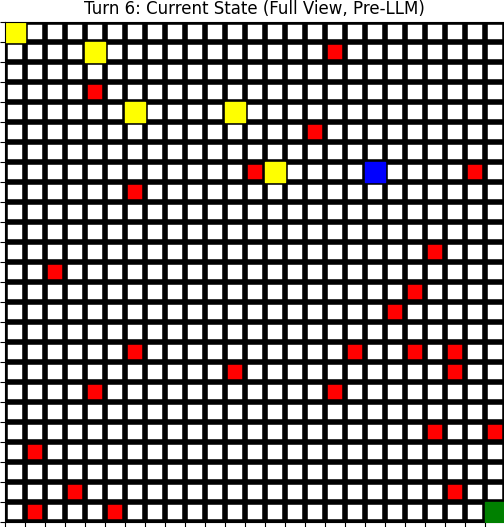} & \includegraphics[width=0.26\linewidth]{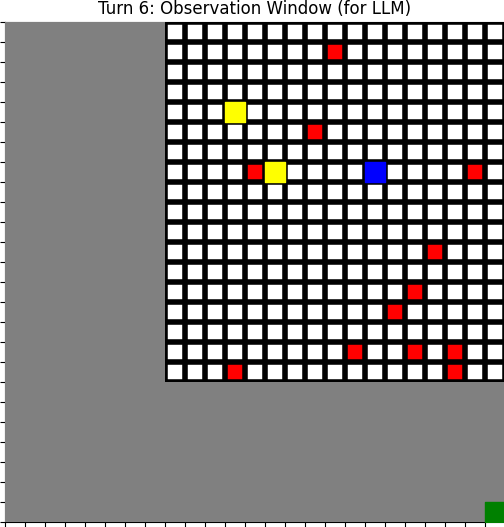} & \includegraphics[width=0.12\linewidth]{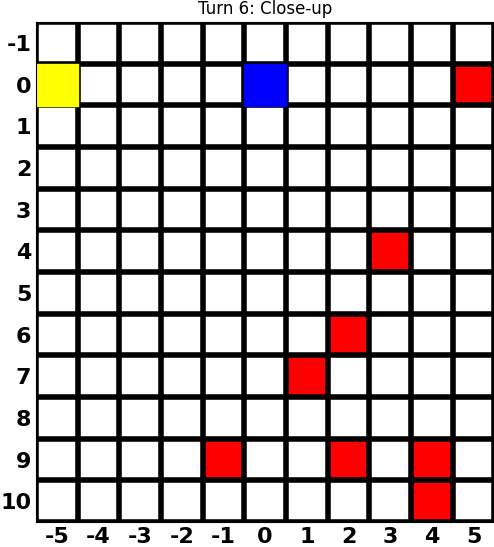} & \includegraphics[width=0.26\linewidth]{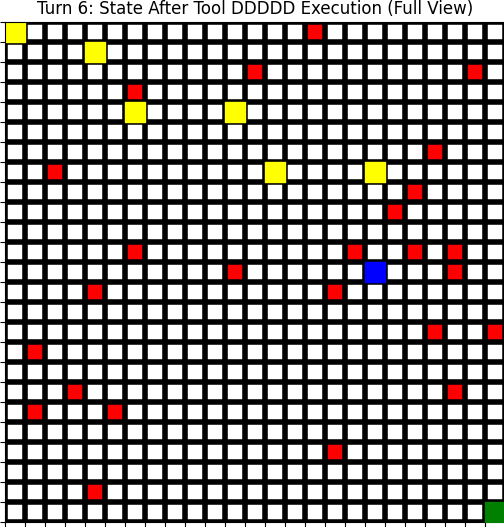} \\ \tiny 7 & \includegraphics[width=0.26\linewidth]{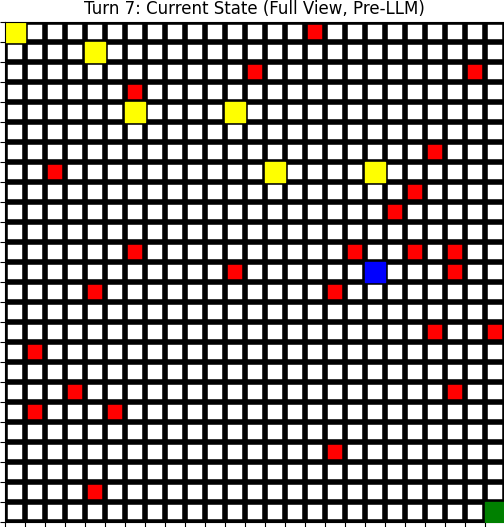} & \includegraphics[width=0.26\linewidth]{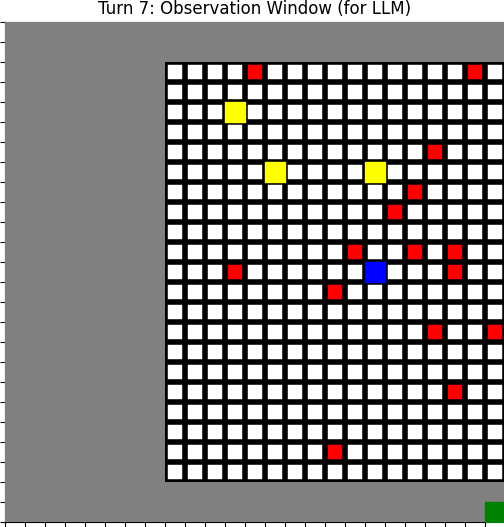} & \includegraphics[width=0.12\linewidth]{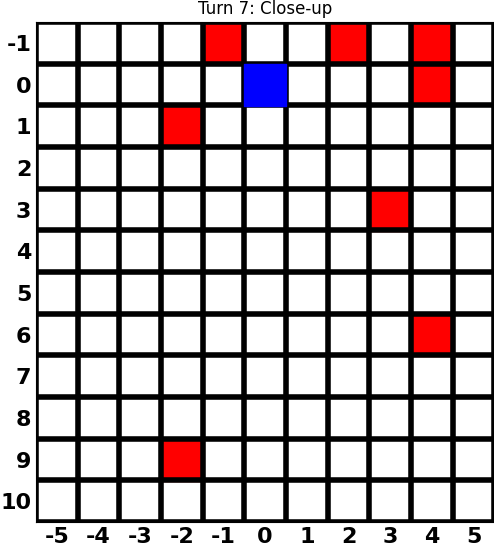} & \includegraphics[width=0.26\linewidth]{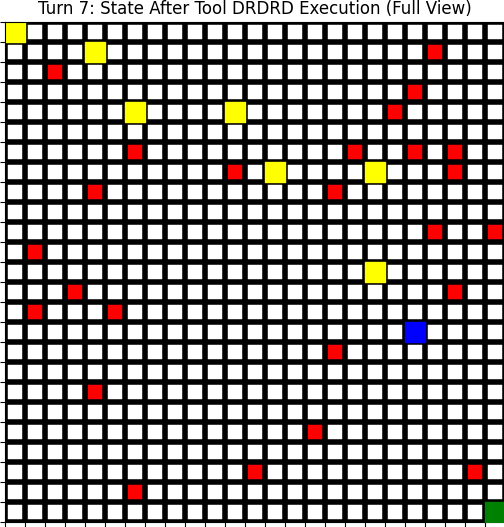} \\ \tiny 8 & \includegraphics[width=0.26\linewidth]{turn_8_full_view_pre_llm.png} & \includegraphics[width=0.26\linewidth]{turn_8_obs_view_for_llm.png} & \includegraphics[width=0.12\linewidth]{turn_8_zoom_view_for_llm.png} & \includegraphics[width=0.26\linewidth]{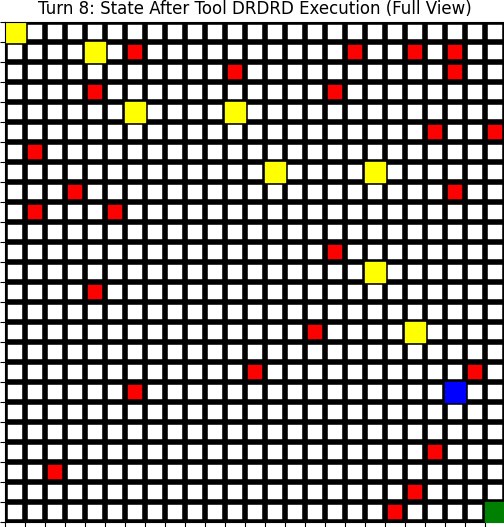} \\ \tiny 9 & \includegraphics[width=0.26\linewidth]{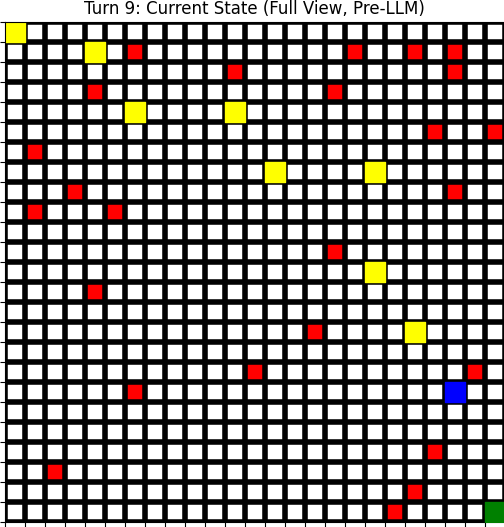} & \includegraphics[width=0.26\linewidth]{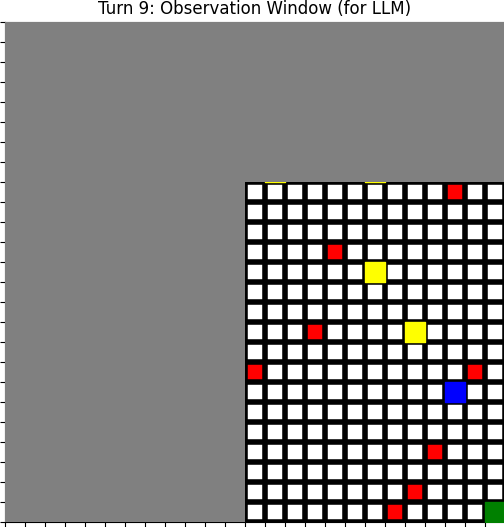} & \includegraphics[width=0.12\linewidth]{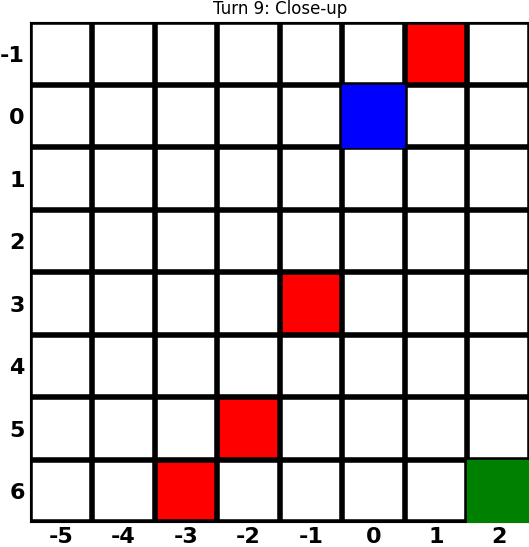} & \includegraphics[width=0.26\linewidth]{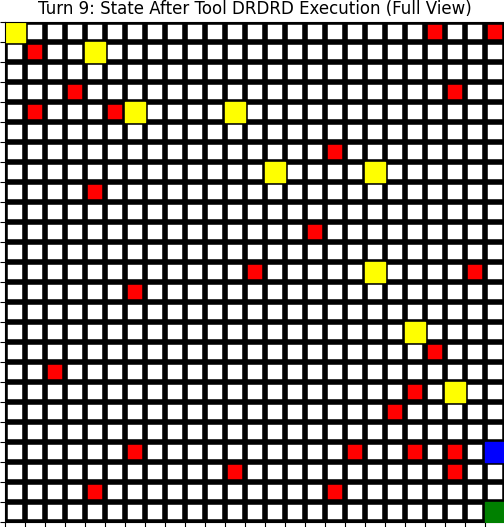} \\ \tiny 10 & \includegraphics[width=0.26\linewidth]{turn_10_full_view_pre_llm.png} & \includegraphics[width=0.26\linewidth]{turn_10_obs_view_for_llm.png} & \includegraphics[width=0.12\linewidth]{turn_10_zoom_view_for_llm.png} & \includegraphics[width=0.26\linewidth]{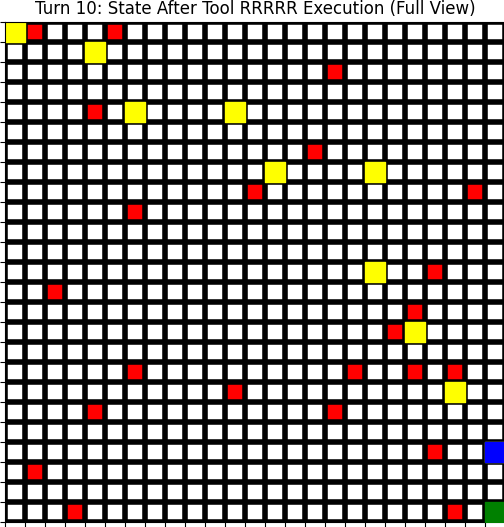}
\end{tabular} \caption{Detailed version of Figure~\ref{fig:results}, turns 6--10 (target reached at turn~10). LLM: Qwen3.6-35B-A3B.} \label{fig:results_detailed_portrait5to10} \end{figure*}

We present in Figures~\ref{fig:results_detailed_portrait1to5}-\ref{fig:results_detailed_portrait5to10} a detailed version of Figure~\ref{fig:results} (the same episode, LLM Qwen3.6-35B-A3B). Each row also shows the agent-centered close-up (``zoom'') next to the observation.

\end{document}